\def\year{2021}\relax
\documentclass[letterpaper]{article} 
\usepackage{aaai21}  
\usepackage{times}  
\usepackage{helvet} 
\usepackage{courier}  
\usepackage[hyphens]{url}  
\usepackage{graphicx} 
\usepackage{natbib}  
\usepackage{caption} 
\usepackage[switch]{lineno} 
\nocopyright
\usepackage{natbib}

\usepackage{listings}
\usepackage{amsmath}
\usepackage{amsfonts}
\usepackage{MnSymbol}

\newtheorem{example}{Example}
\newtheorem{theorem}{Theorem}

\newcommand{\F}{{\cal F}}
\newcommand{\U}{{\cal U}}
\newcommand{\R}{{\cal R}}
\newcommand{\V}{{\cal V}}

\newcommand{\union}{\cup}
\newcommand{\sat}{\models}

\newtheorem{fact}{Fact}
\newcommand{\commentout}[1]{}

\newtheorem{definition}{Definition}
\newcommand{\dfn}{\begin{definition}}
\newcommand{\bbox}{\vrule height7pt width4pt depth1pt}
\newcommand{\edfn}{\bbox\end{definition}}
\newcommand{\thm}{\begin{theorem}}
\newcommand{\ethm}{\end{theorem}}
\newcommand{\prf}{\noindent{\bf Proof:} }
\newcommand{\eprf}{\bbox\vspace{0.1in}}
\newcommand{\xam}{\begin{example}}
\newcommand{\cross}{\times}

\newcommand{\nonblind}[1]{#1}
\newcommand{\longv}[1]{#1}

\title{Equivalent Causal Models}
\title{Equivalent Causal Models}
\author {
    Sander Beckers \\
}

\affiliations{
    Munich Center for Mathematical Philosophy \\
    Ludwig Maximilian University, Munich \\
    srekcebrednas@gmail.com
}

\begin{document}


\maketitle

\begin{abstract}
The aim of this paper is to offer the first systematic exploration and definition of {\em equivalent} causal models in the context where both models are not made up of the same variables. The idea is that two models are equivalent when they agree on all ``essential'' causal information that can be expressed using their common variables. I do so by focussing on the two main features of causal models, namely their structural relations and their functional relations. In particular, I define several relations of causal ancestry and several relations of causal sufficiency, and require that the most general of these relations are preserved across equivalent models.
\end{abstract}

\section{Introduction}

\noindent Causal models are becoming evermore important and their domains of application are growing rapidly, both within science (most notably AI) and in philosophy. As with all forms of modeling, there usually exist many causal models that can be used to describe a given system of interest. Since different models can lead to widely divergent conclusions, it is of paramount importance to have some method to compare models and group together those that are sufficiently similar. To do so we need to be able to establish when two causal models are {\em equivalent}. The aim of this paper is to offer the first systematic exploration and definition of equivalent causal models. 

Of course what makes for a sensible notion of equivalence varies across different contexts. For example, there exists a well-known notion of equivalence that is used in the epistemically limited context where one is learning a causal model from data. However, such Markov-equivalence (as it is called) does not apply in the context where two causal models are fully specified and do not consist of the same variables. Such contexts are widespread in science, but are of particular concern when trying to capture {\em stable}, model-independent, features of the world. Furthermore, exploring equivalence in such an omniscient context offers a firm conceptual starting point for generalizing to contexts that involve approximation and uncertainty in future work. 

Before diving into the details, I spell out the underlying intuition of both the notion to be developed and the method by which I shall do so. The aim is to formalize a suitable definition of causal model-equivalence that is relative to those variables that the models have in common: the idea is that two models are equivalent when they agree on all ``essential'' causal information that can be expressed using their common variables. Obviously this will require motivating what I take to be (in)essential and why I choose to focus merely on the common variables. 

My method focusses on the observation that causal models are characterized using two types of relations: structural relations, i.e., having to do with the causal graph, and functional relations, i.e., having to do with how particular values of variables are determined by values of other variables. I develop two notions of equivalence, one for each type. Causal equivalence will then be defined as being both structurally and functionally equivalent. 

The details are as follows. The most basic structural relation is that of one variable $X$ being an ancestor of another variable $Y$. I generalize this relation along three dimensions. First, I take into account the specific values $x,x'$ and $y,y'$ of the variables that are responsible for the ancestry. Second, I consider ancestry for {\em sets} of variables $\vec{X}$ and $\vec{Y}$. Third, I distinguish between potential ancestors, which are relative to a causal model, and actual ancestors, which are relative to a causal model and a specific context. Structural equivalence is then defined as having identical potential and actual ancestors. 

The most basic functional relation is that of one set of variable values $\vec{x}$ being causally sufficient for another set $\vec{y}$. I first look into spelling out an appropriate notion of causal sufficiency, resulting in direct sufficiency. I then generalize this relation by defining sufficiency as its transitive closure. Functional equivalence is defined as having identical sufficiency relations.

The structure of this paper is as follows: the next section introduces the structural equations framework. I then offer a motivation and clarification of the problem that this paper is concerned with in Section \ref{sec:mot}. Section \ref{sec:ancestors} focusses on the causal structure and develops a notion of structural equivalence. In Section \ref{sec:func} I shift focus to functional equivalence, by considering what makes for a good definition of causal sufficiency. Causal equivalence is then defined as the combination of both forms of equivalence. Section \ref{sec:rel} compares functional equivalence to Halpern's notion of conservative equivalence. Section \ref{sec:con} discusses other related work and reflects on possible future directions.

\section{Structural Equation Models}\label{sec:causal}

This section reviews the definition of causal models as understood in the structural modeling tradition started by \cite{pearl:book}, mostly using the notation and terminology from \cite{halpernbook}. 

\dfn
A signature $\cal S$ is a tuple $(\U,\V,\R)$, where $\U$
is a set of \emph{exogenous} variables, $\V$ is a set 
of \emph{endogenous} variables,
and $\R$ a function that associates with every variable $Y \in  
\U \union \V$ a nonempty set $\R(Y)$ of possible values for $Y$
(i.e., the set of values over which $Y$ {\em ranges}).
If $\vec{X} = (X_1, \ldots, X_n)$, $\R(\vec{X})$ denotes the
crossproduct $\R(X_1) \times \cdots \times \R(X_n)$.
\edfn

Exogenous variables represent factors whose causal origins are outside the scope of the causal model, such as background conditions and noise. 
The values of the endogenous variables, on the other hand, are causally determined by other variables within the model (both endogenous and exogenous). 

\dfn
A \emph{causal model} $M$ is a pair $(\cal S,\F)$, 
where $\cal S$ is a signature and
$\F$ defines a  function that associates with each endogenous
variable $X$ a \emph{structural equation} $F_X$ giving the value of
$X$ in terms of the 
values of other endogenous and exogenous variables. Formally, the equation $F_X$ maps $\R(\U \union \V - \{X\})$  to $\R(X)$,
so $F_X$ determines the value of $X$, 
given the values of all the other variables in $\U \union \V$.  
\edfn

We call a setting
$\vec{u} \in \R(\U)$ of values of exogenous variables a \emph{context}.
The value of $Y$ may depend on the values of only a few other variables.  
$Y$ \emph{depends on $X$}
if there is some context $\vec{u}$ and some setting of the endogenous variables  
other than $Y$ and $X$ such that varying the value
of $X$ in that 
context results in a variation in the value of $Y$; that 
is, there is a setting $\vec{z}$ of the endogenous variables other than $Y$ and
$X$ and values $x$ and $x'$ of $X$ such that $F_Y(x,\vec{z},\vec{u}) \ne
F_Y(x',\vec{z},\vec{u})$. We then say that $X$ is a {\em parent} of $Y$. 
It will prove useful later on to have some notation to capture the details of the foregoing situation: we write $(x,x') \leadsto (y,y')$ and say that $(\vec{u},\vec{z})$ is a {\em witness} to this.

In this paper we restrict attention to \emph{strongly recursive} (or
\emph{acyclic}) models, that is, models where there is a partial order
$\preceq$ on variables such that if $Y$ depends on $X$, then $X \prec Y$.\footnote{All definitions could easily be generalized to {\em recursive} models, i.e., models where the partial order may depend on the context.}
In a strongly recursive model, given a context $\vec{u}$, 
the values of all the endogenous variables are determined (we can just
solve for the value of the variables in the order given
by $\prec$). We often write the
equation for an endogenous variable as $Y = f(\vec{X})$; this denotes
that the value of $Y$ depends only on the values of the variables in
$\vec{X}$, and the connection is given by the function $f$. For example, we might
have $Y = X + 5$.
 
An \emph{intervention} has the form $\vec{X} \gets \vec{x}$, where $\vec{X}$ 
is a set of endogenous variables.  Intuitively, this means that the
values of the variables in $\vec{X}$ are set to the values $\vec{x}$.  
The structural equations define what happens in the presence of 
interventions.  Setting the value of some variables $\vec{X}$ to
$\vec{x}$ in a causal 
model $M = (\cal S,\F)$ results in a new causal model, denoted $M_{\vec{X}
\gets \vec{x}}$, which is identical to $M$, except that $\F$ is
replaced by $\F^{\vec{X} \gets \vec{x}}$: for each variable $Y \notin
  \vec{X}$, $F^{\vec{X} \gets \vec{x}}_Y = F_Y$ (i.e., the equation
  for $Y$ is unchanged), while for
each $X'$ in $\vec{X}$, the equation $F_{X'}$ for $X'$ is replaced by $X' = x'$
(where $x'$ is the value in $\vec{x}$ corresponding to $X'$).

Given a signature $\cal S = (\U,\V,\R)$, an \emph{atomic formula} is a
formula of the form $X = x$, for  $X \in \V$ and $x \in \R(X)$.  
A {\em causal formula (over $\cal S$)\/} is one of the form
$[Y_1 \gets y_1, \ldots, Y_k \gets y_k] \phi$, where
\begin{itemize}
\item $\phi$ is a Boolean combination of atomic formulas,
\item $Y_1, \ldots, Y_k$ are distinct variables in $\V$, and
\item $y_i \in \R(Y_i)$ for each $1 \leq i \leq k$.
\end{itemize}
Such a formula is abbreviated as $[\vec{Y} \gets \vec{y}]\phi$. The special case where $k=0$ is abbreviated as $\phi$.
Intuitively, $[Y_1 \gets y_1, \ldots, Y_k \gets y_k] \phi$ says that $\phi$ would hold if $Y_i$ were set to $y_i$, for $i = 1,\ldots,k$.

A causal formula $\psi$ is true or false in a \emph{causal setting}, which is a causal model given a
context. As usual, we write $(M,\vec{u}) \sat \psi$  if the causal
formula $\psi$ is true in the causal setting $(M,\vec{u})$.
The $\sat$ relation is defined inductively. 
$(M,\vec{u}) \sat X = x$ if the variable $X$ has value $x$
in the unique (since we are dealing with recursive models) solution
to the equations in $M$ in context $\vec{u}$ (i.e., the unique vector
of values that simultaneously satisfies all
equations in $M$ with the variables in $\U$ set to $\vec{u}$).
The truth of conjunctions and negations is defined in the standard way.
Finally, $(M,\vec{u}) \sat [\vec{Y} \gets \vec{y}]\phi$ if 
$(M_{\vec{Y} \gets \vec{y}},\vec{u}) \sat \phi$.

\section{Motivation}\label{sec:mot}

There are many different contexts in which one might consider one causal model $M$ to be equivalent to another model $M'$. A crucial dimension along which we can separate these contexts is by looking at the kind of relation they require to hold between the signatures of both models. 

A first context, that this paper is not directly concerned with, is the one in which both models are required to have identical signatures $(\U,\V,\R)$. It is within this context that most causal search algorithms take place. The output of such an algorithm is usually a partially directed graph that represents the Markov Equivalence Class. 
Markov equivalence means that two models have identical observational probability distributions. Yet two Markov equivalent models only partially agree about the ancestral relations, and need not agree at all about the structural equations. 

This paper is concerned with a second context, namely the one in which we are comparing two causal models only with regard to their common variables (captured by a common part of the signature $(\U,\V,\R)$). In this context, the primary interest of the modeler lies in the causal relations between these common variables. We are looking for a notion of equivalence that is strong enough such that when confronted with a choice between two models $M$ and $M'$ that contain (at least) all of these variables, we can state that if $M$ and $M'$ are equivalent with regards to $(\U,\V,\R)$ then they agree on all the essential causal relations that hold between the common variables and thus one is free to choose either model.

This notion of equivalence matters because we want to be able to draw {\em stable} conclusions about the causal relations between the variables of interest: no matter what further variables we discover to be causally connected to the variables of interest, and no matter what variables that we are not interested in are being marginalized away, as long as the resulting causal model is equivalent to the original one the conclusions remain identical. 

As a consequence, the goal of this paper is to find a much stronger notion of equivalence than any notion that could be useful in the first context, one that gets much closer to identity (which is of course the strongest possible notion of equivalence). In particular, any difference between two equivalent causal models in this strong sense should be entirely explained by the fact that the two models {\em do not have identical signatures}. A fortiori, when applying this notion of equivalence to models with identical signatures, it should reduce to identity. For example, say we have $M_1$ consisting of the equation $E=C$ and $M_2$ consisting of equations $E=D$ and $D=C$. If we are exclusively interested in the variables $E$ and $C$, then any sensible notion of equivalence that allows for different signatures should consider both models to be equivalent with regards to $E$ and $C$. Of course this does not mean that it is impossible to state a causal relation between $C$ and $E$ over which both models disagree: one could simply count the number of edges it takes to get from $C$ to $E$, and in that respect the two models clearly are different. Yet I assume that such ``numerical'' properties are not essential. If such properties were essential, then it seems impossible to come up with any useful notion of equivalence, for presumably one can always zoom in and inject a hitherto ignored intermediate variable along any edge. Likewise, the number of distinct paths going from one variable to another is also deemed to be inessential for the purposes of this paper. I make do with these intuitive considerations regarding numerical properties and refrain from giving a precise definition.
 
Instead of explicitly considering two models that have a partly overlapping signature, I am only going to consider causal models $M = ((\U,\V,\R),\F)$ and $M' = ((\U',\V',\R'),\F')$ such that $\U \subseteq \U'$ and $\V \subseteq \V'$. Given that an equivalence relation is transitive and symmetrical, this restriction is without loss of generality. Further, given that any independent influence of the exogenous variables $\vec{W}=\U' \minus \U$ is lost 
 when only considering $\U$, I assume that the extended model $M'$ is equivalent to $M$ with regards to $(\U,\V,\R)$ {\em for a specific setting $\vec{w} \in \R(\vec{W})$} of all the exogenous variables that are being marginalized. (We call $\vec{w}$ a {\em witness}.) Therefore a setting $(\vec{u}',\vec{v}')$ of $M'$ will be compared to a setting $(\vec{u},\vec{v})$ of $M$, where $\vec{v}$ is the restriction of $\vec{v}'$ to $\V$, $\vec{u}$ is the restriction of $\vec{u}'$ to $\U$, and $\vec{w} \subseteq \vec{u}'$. Any setting of the exogenous variables of $M'$ which does not contain $\vec{w}$ is ignored entirely. 

Lastly a brief word about a third context that one could consider for the relation between the signatures of two models. A natural suggestion is to also compare two models such that we have a mapping $\tau:\R(\U') \cross \R(\V') \rightarrow \R(\U) \cross \R(\V)$ between the values of the variables of both models. Indeed, that was the context \nonblind{investigated in previous work} \citep{BH19_2, BEH19_2}. As such this third context is strictly more general than the second one, for the previous context can be seen as the special case where the mapping is the identity-mapping for the common variables and marginalization for the others, i.e., $\tau(\vec{u}',\vec{v}')=(\vec{u},\vec{v})$ if $\vec{w} \subseteq \vec{u}'$ and undefined otherwise for all $\vec{v}' \in \R(\V')$ and $\vec{u}' \in \R(\U')$. However, in that \nonblind{earlier work the notion of equivalence that we used was} left entirely implicit, and thus other -- more interesting -- notions \nonblind{were} never considered. (This will be discussed in Section \ref{sec:rel}.) So in that respect the current paper is more general. Also, the special mapping I am here considering is of interest in and of itself, and presents enough of a challenge to merit its own treatment. I do intend to generalize the notions of equivalence here developed to more general mappings in future work. 

\section{Structural Equivalence}\label{sec:ancestors}

This section focuses on the {\em causal structure} of the causal model. Several useful concepts are introduced in this regard, which are then combined to suggest a definition of structural equivalence. 

Section \ref{sec:causal} introduced the concept of one variable being a parent of another. It is customary to further generalize this genealogical terminology by taking {\em ancestor} to be the transitive closure of the parent relation, i.e., $X$ is an ancestor of $Y$ if there are variables $V_1$, ..., $V_n$, such that $X$ is a parent of $V_1$, $V_1$ is a parent of $V_2$, ..., and $V_n$ is a parent of $Y$. By treating variables as nodes and drawing directed edges from parents to their children, each strongly recursive causal model induces a unique DAG (Directed Acyclic Graph) that captures the causal structure of the model. We say that $X$ is an {\em ancestor} of $Y$ if $X$ is an ancestor of $Y$ in this DAG.

This graph by itself offers only a rough characterization of all the causal information that the model contains: it tells you which variable can depend on some other variable, but it does not provide details of this dependence. In particular, it does not say which values of some variable depend on what values of some other variable. Such details can be added by refining the standard genealogical relations so that they also take into account the {\em specific values} which instantiate the parent/ancestral relations.

(Throughout the rest of the paper we assume that $M = ((\U,\V,\R),\F)$, $M' = ((\U',\V',\R'),\F')$, $\U \subseteq \U'$, $\V \subseteq \V'$, and $\vec{W} = \U' - \U$. We further assume that $X$ and $Y$ are distinct members of 
$\V$ unless stated otherwise, $x,x'$ are distinct values in $\R(X)$, and likewise for $y,y'$. Similarly, $\vec{X}$ and $\vec{Y}$ are distinct -- possibly overlapping -- subsets of $\V$, and $\vec{x},\vec{x}'$ are non-overlapping sets of values in  $\R(X)$. For brevity, we often leave the variables implicit, writing $\vec{x}$ instead of $\vec{X}=\vec{x}$, for example. All definitions assume that $\vec{X} \subseteq \V$, but this is merely for notational simplicity. The case where $\vec{X} \cap \U \neq \emptyset$ is left implicit.)

\dfn\label{def:par}  We say that {\em $x$ rather than $x'$ is a potential parent of $y$ rather than $y'$} if there exists a witness $(\vec{u},\vec{z})$ for $(x,x')\leadsto (y,y')$.\footnote{Note that potential parenthood is a symmetric relation for the values of the variables: $(x,x') \leadsto (y,y')$ iff $(x',x) \leadsto (y',y)$. This is no longer true for actual parenthood, introduced later on.}

The ancestral generalization is straightforwardly defined as the transitive closure. We use the following notation for {\em potential ancestry}: $(x,x') \leadsto_{p} (y,y')$, where $p=(X,V_1,\ldots,V_n,Y)$ is called a {\em directed path} from $X$ to $Y$.
\edfn

Should potential ancestral relations be preserved among equivalent causal models? As such this relation contains crucial causal information, namely that there's a {\em stepwise connection} between changing some value of $X$ and changing a value of $Y$. But the potential ancestral relation also contains more specific information that is numeric in nature: it states that this {\em connection is given by a single path}. As mentioned before, I consider it a mistake to focus on numerical properties. Indeed, consider the following example.

\begin{example}\label{example1}
Assume the range of $E$ is $\{0,1,2\}$ and all other variables are binary. (Throughout all examples, all variables are binary unless otherwise stated.) The original model $M$ consists simply of the equation $E= 2*C$. The extended model $M'$ consists of $A=C$, $B=C$, and $E= A + B$. Intuitively, I believe that these models should come out as equivalent regarding $C$ and $E$. Yet although $C$ is an ancestor of $E$ in both models, $(1_C,0_C) \leadsto_{p} (2_E,0_E)$ for some path $p$ only in $M$. (The subscripts indicate to which variables the values correspond.)
\end{example}

Still, the essential property that there exists a stepwise connection between $(1_C,0_C)$ and $(2_E,0_E)$ is preserved in $M'$, it is just that this connection now consists of simultaneously considering two paths rather than one: $(1_C,0_C) \leadsto (1_A,1_B; 0_A,0_B) \leadsto (2_E,0_E)$. In other words, we can generalize the potential ancestral relation to allow for steps between sets of variable values rather than steps between single variable values. Graphical relations are not naturally equipped to deal with sets of values, but functional relations are. So in order to make this precise we first express the single-path potential ancestral relations using functional relations, i.e., relations expressed in terms of which formulas are true under certain interventions. 
 
\begin{fact}\label{fact1}  $(x,x') \leadsto (y,y')$ with witness $(\vec{u},\vec{z})$ in $M$ iff  $(M,\vec{u}) \sat [\vec{Z} \gets \vec{z}, X \gets x]Y=y \land [\vec{Z} \gets \vec{z}, X \gets x']Y=y'$.
\end{fact}

As was pointed out in Section \ref{sec:causal}, in strongly recursive models the parents of a variable suffice to determine its value.
Therefore we can always restrict the witness to only include the context and a variable's endogenous parents. In fact, sometimes we can restrict it even further. For example, if the equation for $Y$ is $Y=(A \land B) \lor C$, then $(0_A,0_C)$ suffices to determine that $Y=0$ and $(0_A,1_C)$ suffices to determine that $Y=1$. This allows us to conclude that $(0_C,1_C) \leadsto (0_Y,1_Y)$ with just using $(0_A)$ as a witness.

Generalizing the notion of a witness so that it need not consist of values for all other variables allows for comparing potential parenthood across models with different signatures. We make this precise by defining what it means for a set of variable values to be {\em directly sufficient} for another set of variable values. Informally, a set is sufficient precisely when the outcome is independent of the values of variables not in the set.

\dfn\label{def:dirstrsuf2} We say that {\em $\vec{x}$ is directly sufficient for $\vec{y}$} if for all settings $\vec{z}$ of the other endogenous variables and all contexts $\vec{u}$ we have that $(M,\vec{u})\sat [\vec{X} \gets \vec{x},\vec{Z} \gets \vec{z}] \vec{Y}=\vec{y}$. 
\edfn

Now that we have identified the crucial functional relation out of which the potential parent relation is built up, we can generalize this parent relation to sets of variable values.

\dfn\label{def:jointpar} We say that {\em $\vec{x}$ rather than $\vec{x}'$ are potential joint parents of $\vec{y}$ rather than $\vec{y}'$} if there is a set $\vec{Z} \subseteq \V \cup \U$, and a setting $\vec{z} \in \R(\vec{Z})$ such that 
\begin{itemize}
	\item $(\vec{z},\vec{x})$ is directly sufficient for $\vec{y}$,
	\item $(\vec{z},\vec{x}')$ is directly sufficient for $\vec{y}'$, and 
	\item $\vec{X}$ is minimal (i.e., at least one of the previous statements is false when considering any strict subset of $\vec{X}$).
\end{itemize}
We notate this as $(\vec{x},\vec{x}') \leadsto (\vec{y},\vec{y}')$ and say that $\vec{z}$ is a {\em witness}.
\edfn

It is trivial to see that if restricted to singletons, Definition \ref{def:jointpar} reduces to Definition \ref{def:par}. We demand $\vec{X}$ to be minimal in order to avoid calling redundant elements parents. (For example, if we have as structural equation $Y=A$, then we do not want to call $(1_A,1_B)$ joint parents of $Y=1$: $B=1$ is entirely redundant here.) We demand that $\vec{X}$ and $\vec{Y}$ are non-identical in order to avoid self-parenthood. But it would be too much to demand that their intersection is empty. To see why, we consider a slight variant on Example \ref{example1}.

\begin{example}\label{example2}
All variables are as before. The original model $M$ consists simply of the equation $E= 2*C$. The extended model $M'$ consists of $A=C$ and $E= A + C$. As before, I believe that these models should come out as equivalent regarding $C$ and $E$. But in order to establish that $(1_C,0_C) \leadsto_{p} (2_E,0_E)$ for some path $p$ in $M'$, we need to break up the single step into $(1_C,0_C) \leadsto (1_C,1_A; 0_C,0_A) \leadsto (2_E,0_E)$.
\end{example}

Now that we have generalized the potential parenthood relation along the dimension of sets of variables (which is already built into functional relations), we can take the transitive closure to further generalize this relation along the dimension that is characteristic of the ancestral relation (but is lacking from many functional relations).

\dfn\label{def:jointanc} We say that {\em $\vec{x}$ rather than $\vec{x}'$ are potential joint ancestors of $\vec{y}$ rather than $\vec{y}'$} if there exist $\vec{V}_1$, ..., $\vec{V}_n$ and settings $\vec{v}_i,\vec{v}_i' \in \R(\vec{V}_i)$ for each $i \in \{1,\ldots,n-1\}$ such that $(\vec{v}_i,\vec{v}_i') \leadsto (\vec{v}_{i+1},\vec{v}_{i+1}')$, $(\vec{x},\vec{x}') \leadsto (\vec{v}_1,\vec{v}_1')$ and $(\vec{v}_n,\vec{v}_n') \leadsto (\vec{y},\vec{y}')$. We call the tuple $n=(\vec{X},\vec{V}_i,\ldots,\vec{V}_n,\vec{Y})$ a {\em network} from $\vec{X}$ to $\vec{Y}$, and say that {\em $\vec{x}$ rather than $\vec{x}'$ are potential joint ancestors of $\vec{y}$ rather than $\vec{y}'$ along $n$}. We notate this as $(\vec{x},\vec{x}') \leadsto_{n} (\vec{y},\vec{y}')$. 
\edfn

Preservation of potential joint ancestors deals with Examples \ref{example1} and \ref{example2}, but it still does not suffice for our strong notion of equivalence, because it entirely ignores that much causal information is {\em context-dependent}. The above defined ancestral relations are all {\em general properties of the causal model $M$}, in the sense that they do not make any reference to an actual context $\vec{u}$. Yet causal models also capture causal information regarding specific contexts. Indeed, it is such information that is at stake when speaking of actual causation, an important concept that has benefitted greatly from the advent of causal models.
A good notion of structural equivalence should take this into account as well. Therefore we also define a context-specific counterpart of the potential ancestral relations.

\dfn\label{def:actpar} We say that {\em $x$ rather than $x'$ is an actual parent of $y$ rather than $y'$ w.r.t.~$(M,\vec{u})$} if $(x,x') \leadsto (y,y')$ with a witness $(\vec{u},\vec{z})$ such that $(M,\vec{u}) \sat \vec{Z} =\vec{z} \land X=x$. We notate this as $(x,x') \leadsto^{\vec{u}} (y,y')$, and say that $(\vec{u},\vec{z})$ is a {\em witness} to this.
\edfn

Both the generalization of actual parenthood to sets of variable values and the analogous ancestral counterpart are defined straightforwardly. I only explicitly give the combination of both generalizations, which is the context-relative counterpart of Definition \ref{def:jointanc}.

\dfn\label{def:jointactanc}
We say that {\em $\vec{x}$ rather than $\vec{x}'$ are actual joint ancestors of $\vec{y}$ rather than $\vec{y}'$ w.r.t.~$(M,\vec{u})$} if $(\vec{x},\vec{x}') \leadsto_{n} (\vec{y},\vec{y}')$ for some $n$ such that the exogenous parts of all the witnesses are contained in $\vec{u}$ and all the endogenous parts $\vec{z}$ are such that $(M,\vec{u}) \sat \vec{Z} =\vec{z} \land \vec{X}=\vec{x}$. We notate this as $(\vec{x},\vec{x}') \leadsto^{\vec{u}}_{n} (\vec{y},\vec{y}')$.
\edfn

The following example illustrates why preservation of actual joint ancestors should be combined with the preservation of potential joint ancestors to get structural equivalence.

\begin{example}\label{example4}
Model $M$ contains $E = B \lor D$, whereas model $M'$ contains $E = (A \land \lnot C) \lor B \lor D$. Both models contain $A=C$, $B=C$, and $D=\lnot C$. Note that both models have identical signatures. It is easy to verify that for each context both models will have the same actual joint ancestors. Yet they disagree on the potential ancestors, for $(1_A,0_A) \leadsto (1_E,0_E)$ in $M'$ (with witness $(0_C,0_B,0_D)$) and not so in $M$. The two models are indeed different, for they disagree on whether $A$ is an ancestor of $E$, and thus do not even have the same DAG.
\end{example}

We are finally ready to define what it means for the structural causal information to be preserved across causal models. 

\dfn\label{def:struceq} $M$ is {\em structurally equivalent} to $M'$ if there exists a setting $\vec{w} \in \R(\vec{W})$ such that for all $\vec{X},\vec{Y}$ and  $\vec{x}$, $\vec{y}$ the following two conditions hold:

$(\vec{x},\vec{x}') \leadsto_{n} (\vec{y},\vec{y}')$ for some $n$ in $M$ iff $(\vec{x},\vec{x}') \leadsto_{m} (\vec{y},\vec{y}')$ for some $m$ in $M'$.

For all $\vec{u} \in \R(\U)$ it holds that

$(\vec{x},\vec{x}') \leadsto^{\vec{u}}_{n} (\vec{y},\vec{y}')$ for some $n$ in $M$ iff $(\vec{x},\vec{x}') \leadsto^{(\vec{w},\vec{u})}_{m} (\vec{y},\vec{y}')$ for some $m$ in $M'$.

We call $\vec{w}$ the {\em witness} for $M$ being structurally equivalent to $M'$. Also, we say that $M'$ is a {\em structural extension} of $M$, and $M$ is a {\em structural reduction} of $M'$.
\edfn

The following variation on Example \ref{example4} illustrates that structural equivalence does not suffice to ensure that equivalence of models with identical signatures reduces to identity.

\begin{example}\label{example_struc}
In $M$ we have $E = (A \land \lnot C \land F) \lor B \lor D \lor (G \land F)$, whereas in $M'$ we have $E = (A \land \lnot C \land G) \lor B \lor D \lor (G \land F)$. Both models contain $A=C$, $B=C$, $F=C$, $G=C$, and $D=\lnot C$. The reader can confirm that in addition to agreeing on all actual joint ancestral relations, as was the case with Example \ref{example4}, they also agree on all potential joint ancestors, and are thus structurally equivalent. (Note that they sometimes disagree about the witnesses for the potential relations: $(1_F,0_F) \leadsto (1_E,0_E)$ with witness $(1_A,0_C,0_B,0_D,0_G)$ only in $M$, and likewise with $F$ and $G$ reversed for $M'$.) Intuitively the models are not equivalent, because they disagree on whether it is $F$ or $G$ that can contribute to $E=1$ together with $A$ and $C$.
\end{example}

The next section develops several notions of equivalence which do give the desired outcome for cases with identical signature, and argues for the superiority of one of them.

\section{Functional Equivalence}\label{sec:func}

Now we switch attention to the basic functional relations that are captured by causal models, meaning those relations capturing how the values of variables are determined by the values of other variables. Recall that if $\vec{x}$ is directly sufficient for $\vec{y}$ (Def. \ref{def:dirstrsuf2}), this means  precisely that $\vec{x}$ fully determines the variables $\vec{Y}$ taking on the values $\vec{y}$.  So the functional relations between a set of variables are specified completely by stating all relations of direct sufficiency that hold between their values. A naive suggestion might therefore be to define functional equivalence as having identical relations of direct sufficiency for all sets only containing variables that are common to both models.  

However, that would give an overly strong notion of equivalence, for it means that the only way of extending a model is by adding children to childless variables. The reason is that the moment one adds a new parent $A$ to a set of existing parents $\vec{X}$ of a variable $Y$ then, by definition of being a parent, at least one setting $\vec{x}$ will cease being directly sufficient for some $y$. To illustrate, even the trivial models we considered as a sanity check in Section \ref{sec:mot} would not be considered equivalent. Recall that $M_1$ is made up of $E=D$ and $D=C$, and $M_2$ consists simply of $E=C$. Given that $C=c$ is directly sufficient for $E=e$ only in the latter, these models would not come out as equivalent for the variables $\{E,C\}$. 

The problem is that, as the name suggests, {\em direct} sufficiency focusses on parent-children relations, and is thus a numerical property. Instead, we generalize the notion of direct sufficiency so that it also captures sufficiency for descendants that are not children. Recall that we generalized ancestral relations by incorporating a feature of functional relations, namely that it also applies to sets of variable values and not just to single variables. Here we make the symmetrical counterpart of that move: we generalize the functional relation of direct sufficiency by incorporating a feature of ancestral relations, namely its transitivity.

\dfn\label{def:strsuf2} We say that {\em $\vec{x}$ is sufficient for $\vec{y}$} if there exist $\vec{w}_0$,...,$\vec{w}_n$ such that for each $i$, $\vec{w}_i$ is directly sufficient for $\vec{w}_{i+1}$, where $\vec{x}=\vec{w}_0$ and $\vec{y}=\vec{w}_n$.
\edfn

This gives rise to our notion of functional equivalence.

\dfn\label{def:streq} $M$ is {\em functionally equivalent} to $M'$ if there exists a setting $\vec{w} \in \R(\vec{W})$ such that for all $\vec{X} \subseteq (\V \cup \U)$, $\vec{Y} \in \V$ and $\vec{x},\vec{y}$, it holds that $\vec{x}$ is sufficient for $\vec{y}$ in $M$ iff $(\vec{w},\vec{x})$ is sufficient for $\vec{y}$ in $M'$.
\edfn

It is easy to see that functional equivalence deals with Example \ref{example_struc}. For example, $(1_A,0_C,1_F,0_B,0_D,0_G)$ is sufficient for $E=1$ in $M$ but not in $M'$.  

Nonetheless, the following example illustrates why functional equivalence by itself does not suffice to guarantee structural equivalence.

\begin{example}\label{ex:final}
$M$ consists simply of $Y = D$ and $X = D$. $M'$ on the other hand consists of $Y = (V \land X) \lor D $, $V = D$, and $X = D$. It is easy to verify that the models are functionally equivalent. Yet both models disagree about whether $X$ is an ancestor of $Y$, and thus are not structurally equivalent. 
\end{example}

We therefore define causal equivalence as the combination of structural and functional equivalence.

\dfn\label{def:causeq} $M$ is {\em causally equivalent} to $M'$ if there exists a witness $\vec{w} \in \R(\vec{W})$ such that the models are both structurally and functionally equivalent w.r.t. $\vec{w}$.
\edfn

\section{Conservative Equivalence}\label{sec:rel}

We have come across two sensible notions of causal sufficiency already (Defs. \ref{def:dirstrsuf2} and \ref{def:strsuf2}). The following definition also offers a very natural notion of causal sufficiency.

\dfn\label{def:dirweasuf} We say that {\em $\vec{x}$ is weakly sufficient for $\vec{y}$ w.r.t.~$(M,\vec{u})$} if $(M,\vec{u})\sat [\vec{X} \gets \vec{x}] \vec{Y}=\vec{y}$. 
\edfn

Direct sufficiency required $\vec{x}$ to guarantee $\vec{y}$ regardless of the context and of the values of the other variables. Weak sufficiency on the other hand merely requires that  $\vec{x}$ helps in getting $\vec{y}$ (even if only by not preventing it), but it does not guarantee it: it relies on the other variables to play nice and take on their usual values. Note that direct sufficiency implies sufficiency, and sufficiency implies weak sufficiency. Therefore one might wonder whether defining equivalence using weak sufficiency could offer a plausible alternative to functional equivalence.

\cite{halpern_cons}  considers precisely this definition of equivalence. Informally, it states that both models agree on all formulas consisting solely of variables that they have in common.
\footnote{Halpern formulates his definition differently, but he shows that both formulations are equivalent. This definition is also slightly more general than his, because he only considers cases in which both models have identical exogenous variables..}

\dfn\label{def:cons} $M$ is {\em conservatively equivalent} to $M'$ if there exists a setting $\vec{w} \in \R(\vec{W})$ such  that for all $\vec{X},\vec{Y}$, all $\vec{x},\vec{y}$, and all $\vec{u}$, it holds that $(\vec{u},\vec{x})$ is weakly sufficient for $\vec{y}$ in $M$ iff $(\vec{w},\vec{u},\vec{x})$ is weakly sufficient for $\vec{y}$ in $M'$.
\edfn

Halpern calls the extended model a ``conservative'' extension precisely because he takes it to be intuitive that the larger model preserves all relevant causal information of the smaller model (as far as their mutual variables go of course). 
Conservative equivalence captures important features that are ignored by structural equivalence. In particular, it meets our demand that it reduces to identity when applied to models with identical signatures (and thus also deals with Example \ref{example_struc}). 

As one would expect, functional equivalence is stronger than conservative equivalence.

\thm\label{thm:strongtoweak} Given $M$, $M'$, and a witness $\vec{w} \in \R(\vec{W})$, it holds that if $M$ and $M'$ are functionally equivalent then they are conservatively equivalent.
\ethm

\longv{
\prf
Assume $M$ and $M'$ are functionally equivalent. Lemma 4.3 from \cite{halpern_cons} implies that it suffices to show the following: for any context $\vec{u}$, variable $X \in \V$, and setting $\vec{z}$ with $\vec{Z}=\V - \{X\}$, it holds that $(M,\vec{u}) \sat [\vec{Z} \gets \vec{z}]X=x$ iff $(M',(\vec{w},\vec{u})) \sat [\vec{Z} \gets \vec{z}]X=x$. 

Say $(M,\vec{u}) \sat [\vec{Z} \gets \vec{z}]X=x^*$. Since $\vec{Z}$ contains all of $X$'s parents, $(\vec{u},\vec{z})$ is directly sufficient for $x^*$ in $M$. Because $M$ and $M'$ are functionally equivalent, this implies that $(\vec{w},\vec{u},\vec{z})$ is sufficient for $x^*$ in $M'$, from which it follows that $(M',(\vec{w},\vec{u})) \sat [\vec{Z} \gets \vec{z}]X=x^*$. If $x=x^*$, the above equivalence holds because both sides are true. If $x \neq x^*$, it holds because both sides are false.
\eprf
}

The following interesting result illustrates that the dichotomy between structural and functional relations is not strict: conservative equivalence (and thus also functional equivalence) is not entirely independent from structural equivalence. Concretely, the ancestor relation is preserved when moving from a causal model to a conservative extension. (Example \ref{ex:final} illustrates that the reverse is not true.)

\thm\label{thm:graph}
Given conservatively equivalent models $M$ and $M'$, for any variables $X, Y \in \V$, we have that if $X$ is an ancestor of $Y$ in $M$ then $X$ is an ancestor of $Y$ in $M'$. In other words, the DAG of $M$ is a sub-DAG of $M'$ when ignoring all the variables not in $\V$.
\ethm

\longv{
\prf 
Assume $X$ is an ancestor of $Y$ in $M$. Because of the transitivity of the ancestral relation, we may assume that 
$X$ is a parent of $Y$ 
 in $M$, i.e., there exist $x,x',y,y'$ and a witness $(\vec{u},\vec{z})$ with $\vec{Z}=\V - \{X,Y\}$ so that $(x,x') \leadsto (y,y')$. By Fact \ref{fact1} this means that $(M,\vec{u}) \sat [\vec{Z} \gets \vec{z}, X \gets x]Y=y \land [\vec{Z} \gets \vec{z}, X \gets x']Y=y'$. Because $M$ and $M'$ are conservatively equivalent, we also have $(M',(\vec{w},\vec{u})) \sat [\vec{Z} \gets \vec{z}, X \gets x]Y=y \land [\vec{Z} \gets \vec{z}, X \gets x']Y=y'$. 

Say $\vec{A}$ are the parents of $Y$ in $M'$ excluding $X$. Thus any setting of $(\vec{A},X)$ combined with the context is directly sufficient for some value of $Y$. If all variables in $\vec{A}$ get the same values under both interventions, i.e., there exists a setting $\vec{a} \in \R(\vec{A})$ such that $(M',(\vec{w},\vec{u})) \sat [\vec{Z} \gets \vec{z}, X \gets x]\vec{A}=\vec{a} \land [\vec{Z} \gets \vec{z}, X \gets x']\vec{A}=\vec{a}$, then $(\vec{w},\vec{u},\vec{a})$ is a witness for the fact that $X$ is also a parent of $Y$ in $M'$.

If they do not get the same values then there exists an $A_1 \in \vec{A}$ which gets different values $a_1$ and $a_1'$ under the respective interventions. In other words, $(M',(\vec{w},\vec{u})) \sat [\vec{Z} \gets \vec{z}, X \gets x]A_1=a_1 \land [\vec{Z} \gets \vec{z}, X \gets x']A_1=a_1'$. 

We are thus in the same situation for $A_1$, a parent of $Y$, as we were for $Y$. Therefore we can repeat the reasoning from above by looking at the parents of $A_1$ in $M'$ excluding $X$. Given strong recursivity, eventually we have to find a variable $A_i$ such that $X$ is a parent of $A_i$, $A_i$ is a parent of $A_{i-1}$, ..., and $A_1$ is a parent of $Y$.
\eprf
}

Despite this result and the intuitive appeal of the idea that equivalence comes down to comparing formulas made up solely of common variables, the following example shows that conservative equivalence is too weak to preserve all causal information that I would consider as relevant, and therefore is not suitable to replace functional equivalence.

\begin{example}\label{ex:cons} The model $M'$ contains the following equations: $E= B \lor C \lor F$, $B=A \land \lnot C$, $C=D$, $A=\lnot D$, and $F=\lnot C$. If we remove variable $B$ we get the model $M$ where the equation for $E$ has become $E=A \lor C \lor F$. The reader can verify that the models are both conservatively and structurally equivalent. Yet the role of $A$ in both models is intuitively quite different: if $C=1$, then $A=1$ and $C=1$ fulfill entirely symmetrical roles in bringing about $E=1$ according to $M$, which is clearly not the case according to $M'$. Functional equivalence formalizes this intuition: $A=1$ is sufficient for $E=1$ in $M$ and not in $M'$.
\end{example}

\section{Other Related Work and Future Directions}\label{sec:con}

\cite{bongers16} offer a definition of ``equivalence after marginalization'' that generalizes Halpern's notion of conservative equivalence to cyclic and non-deterministic models. Their work is highly complementary to mine: contrary to them, I restrict myself to acyclic deterministic models, and contrary to me, they restrict themselves to using just this one type of equivalence (i.e., just as Halpern, they only consider Definition \ref{def:cons}).

The naive suggestion mentioned earlier of defining functional equivalence using direct sufficiency 
 is present implicitly in \citep{BH19_2}. (In the terminology we used there, it comes down to stating that $M$ is a constructive abstraction of $M'$ for the identity-marginalization mapping discussed in Section \ref{sec:mot}). \cite{Rub17} also give such a definition (see their Theorem 9). However, the focus of both articles lies with the third context rather than the second context, as mentioned in Section \ref{sec:mot}. As a result, neither of them discuss the issue of causal equivalence, nor do they consider any of the other forms of equivalence here introduced. 
 
In future work I intend to generalize my approach to the third context by combining my notions of equivalence with the various notions of abstraction from  \citep{BH19_2}. Simply put, the idea would be to require for some causal relations that they hold between $\vec{x}$ and $\vec{y}$ in $M$ iff they hold between $\tau(\vec{x})$ and $\tau(\vec{y})$ in $M'$, where $\tau$ is an ``abstraction'' mapping. (What makes this non-trivial, is the fact that $\tau$ is defined as mapping a setting $\vec{v} \in \R(\V)$ to some $\vec{v}' \in \R(\V')$, and thus one has to come up with a sensible generalization to random settings $\vec{x}$ and $\vec{y}$.) As an additional step, I can make use of the work from  \citep{BEH19_2}  to define a notion of approximate equivalence.
 
Another promising extension concerns the topic of actual causation, i.e., the causal relation that holds between particular events, as opposed to types of events. I intend to use the current approach to analyse existing -- and to propose novel -- definitions of actual causation by assuming that a good definition ought to be stable across equivalent models: whether $\vec{x}$ causes $\vec{y}$ in some context $\vec{u}$ should not depend on which of two equivalent models one uses. I take the idea of such an analysis from \cite{gallow19}, whose work offered the impetus for the current project. He offers a very strong sufficient condition for equivalence, but does not develop a notion of equivalence. (The interested reader can easily verify that his sufficient condition indeed suffices to meet the criteria of my definition of causal equivalence.) 
 
The notion of actual joint ancestry (Definition \ref{def:jointactanc}) will be central to this extension. It was inspired by a similar -- but subtly different -- one from \cite{gallow19}, who presents it as the basis for a stable definition of actual causation.\footnote{For the interested reader: here is an example for which his definition offers a different verdict. The equations are $E= (A \lor B) \land C$, and $B=A\lor C$, and we consider a context in which $A=1$ and $C=1$. Then $(1_A,1_C; 0_A,0_C) \leadsto^{\vec{u}} (1_A,1_B; 0_A,0_B) \leadsto^{\vec{u}} (1_E,0_E)$, and thus $(A=1,C=1)$ are actual joint ancestors of $E=1$. Gallow's definition involves a minimality condition applied to the entire network, as opposed to applied to each step between parents and children. Since also $(1_C,0_C) \leadsto^{\vec{u}}  (1_E,0_E)$, $(A=1,C=1)$ is not minimal and therefore he does not consider it an actual cause.}\footnote{His full definition adds to this a focus on the default/deviant distinction between values of variables, which is fairly common in work on actual causation. This distinction and his use of it could easily be added to my analysis, and thus stands orthogonal to it.} Therefore the current paper can be seen as offering a partial formal vindication of his approach (modulo the subtle differences between the definitions). But I believe there is still room for improvement. For example, actual joint ancestry is (by definition) a transitive relation, whereas nowadays there is a broad consensus that actual causation is not transitive. For a more concrete and intuitive objection, consider a model with equation $E = (C \land B) \lor  (\lnot C \land A)$, and a context such that $C=1$, $B=1$, and $A=1$. Then $(1_A,1_C; 0_A,0_C) \leadsto^{\vec{u}} (1_E,0_E)$.  However, it seems strange to consider $A=1$ to be a part of an actual cause of $E=1$, given that the disjunct it appears in was false.
\commentout{
Difference 1: effects for him are always singletons. But not sure that this matters, because he doesn't require there to be a sequence, and so can consider sets in the effect by looking at singletons in parallel.
Difference 2: he allows that, with the exception of the end variable E, all contrast values are the actual values. Of course I also allow that, but then I don't consider those variables part of the ancestors: they're part of the witness. (Even if I were to allow it, it would be impossible due to minimality.) Still, he only needs these variables due to this focus on default/deviant: otherwise it is redundant. 
Difference 3: he requires minimality for the entire network, not per parent-children step.
}

To sum up, I investigated when two causal models sharing a subset of variables ought to be considered equivalent. I proposed an answer by looking at the two fundamental features of a causal model, namely its structural and its functional relations. Although the discussion has made clear that this definition satisfies certain intuitive properties, I do not take it to offer the last word. There may very well be further intricacies that the current definition overlooks, which could lead to more refined notions of equivalence.

\subsubsection{Acknowledgments.}

Many thanks to Joe Halpern for comments on an earlier version of this paper. This research was made possible by funding from the Alexander von Humboldt Foundation.

\bibliographystyle{spbasic} 
\bibliography{joe,allpapers}

\end{document}